# Methods for Segmentation and Classification of Digital Microscopy Tissue Images

Quoc Dang Vu[1], Simon Graham[2], Minh Nguyen Nhat To[1], Muhammad Shaban[2], Talha Qaiser[2], Navid Alemi Koohbanani[2], Syed Ali Khurram[3], Tahsin Kurc[4], Keyvan Farahani[5], Tianhao Zhao[4,6], Rajarsi Gupta[4,6], Jin Tae Kwak[1], Nasir Rajpoot[2], Joel Saltz[4]

[1]Department of Computer Science and Engineering, Sejong University, Seoul, Korea
[2]Department of Computer Science, University of Warwick, UK
[3]The University of Sheffield, Sheffield, UK
[4]Department of Biomedical Informatics, Stony Brook University, Stony Brook, USA
[5]Cancer Imaging Program, National Cancer Institute, National Institutes of Health, USA
[6]Department of Pathology, Stony Brook University, Stony Brook, USA

**Abstract**. High-resolution microscopy images of tissue specimens provide detailed information about the morphology of normal and diseased tissue. Image analysis of tissue morphology can help cancer researchers develop a better understanding of cancer biology. Segmentation of nuclei and classification of tissue images are two common tasks in tissue image analysis. Development of accurate and efficient algorithms for these tasks is a challenging problem because of the complexity of tissue morphology and tumor heterogeneity. In this paper we present two computer algorithms; one designed for segmentation of nuclei and the other for classification of whole slide tissue images. The segmentation algorithm implements a multiscale deep residual aggregation network to accurately segment nuclear material and then separate clumped nuclei into individual nuclei. The classification algorithm initially carries out patch-level classification via a deep learning method, then patch-level statistical and morphological features are used as input to a random forest regression model for whole slide image classification. The segmentation and classification algorithms were evaluated in the MICCAI 2017 Digital Pathology challenge. The segmentation algorithm achieved an accuracy score of 0.78. The classification algorithm achieved an accuracy score of 0.81. These scores were the highest in the challenge.

## 1. INTRODUCTION

Cancer causes changes in tissue at the sub-cellular scale. Pathologists examine a tissue specimen under a powerful microscope to look for abnormalities which indicate cancer. This manual process has traditionally been the de facto standard for diagnosis and grading of cancer tumors. While it continues to be widely applied in clinical settings, manual examination of tissue is a subjective, qualitative analysis and is not scalable to translational and clinical research studies involving hundreds or thousands of tissue specimens. A quantitative analysis of normal and tumor tissue, on the other hand, can provide novel insights into observed and latent sub-cellular tissue characteristics and can lead to a better understanding of mechanisms underlying cancer onset and progression (Madabhushi 2009; Cooper et al. 2018; Chennubhotla et al. 2017).

Technology for whole slide tissue imaging has advanced significantly over the past 20 years. We refer to digital images of tissue specimens that are stained and fixated on a glass slide as *whole slide tissue images (WSIs)*. Highly detailed images of tissue, ranging from 20,000x20,000 pixels to over 100,000x100,000 pixels in resolution, can be captured rapidly with the state-of-the-art tissue image scanners. Improvements in storage and computational technology also have made it possible to store and analyze WSIs. These advances provide significant opportunities for quantitative analysis of tissue morphology. Quantitative analyses not only allow researchers to assemble a more detailed description of tumor structure and heterogeneity but also enable studies with large numbers of tissue samples. As the capacity to rapidly generate large quantities of WSI data has become feasible and is more widely deployed, there is an increasing need for reliable and efficient (semi-)automated computer methods to complement the traditional manual examination of tissue.

The two most common tasks in whole slide tissue image analysis are the segmentation of microscopic structures, like nuclei and cells, in tumor and non-tumor regions and the classification of image regions and whole images. Computerized detection and segmentation of nuclei is one of the core operations in histopathology image analysis. This operation is crucial to extracting, mining and interpreting sub-cellular morphologic information from digital slide images. Cancer nuclei differ from other nuclei in many ways and influence tissue in a variety of ways. Accurate quantitative characterizations of the shape, size and texture properties of nuclei are key components of the study of the tumor systems biology and the complex patterns of interaction between tumor cells and other cells. Image classification, carried out with or without segmentation, assigns a class label to an image region or an image. It is a key step in computing a categorization via imaging features of patients into groups for cohort selection and correlation analysis. Methods for segmentation and classification have been proposed by several research projects (Xing and Yang 2016; Peikari and Martel, 2016; Zheng et al. 2017; Gurcan *et al.*, 2009; Senaras, 2018; Chen et al, 2017; Wang et al. 2016; Xie et al. 2015; Sirinukunwattana et al. 2016; Ghaznavi *et al.*, 2013; Manivannan et al. 2016; Al-Milaji et al. 2017; Xu et al. 2015; Graham and Rajpoot, 2018). Xing and Yang (Xing and Yang 2016) provide a good review of segmentation algorithms for histopathology images. A CNN algorithm was developed by Zheng et al. (Zheng et al. 2017) to analyze histopathology images for extraction and characterizations of distribution of nuclei in images of tissue specimens. A method based on ensembles of support vector machines for detection and classification of cellular patterns in tissue images was proposed by Manivannan et al. (Manivannan et al. 2016). Al-Milaji et al. developed a CNN-based approach to classify tissue regions into stromal and epithelial in images of Hematoxylin and Eosin (H&E) stained tissues (Al-Milaji et al. 2017). Xu et al. (Xu et al. 2015) used a pre-trained CNN model to extract features on patches. These features are aggregated to classify whole slide tissue images. A method that learns class-specific dictionaries for classification of histopathology images was proposed by Vu et al. (Vu et al. 2016). Kahya et al. (Kayha et al. 2017) employed support vector machines for classification of breast cancer histopathology images. Their method employs sparse support vector machines and Wilcoxon rank sum test to assign and assess weights of imaging features. Peikari et al. (Peikari et al. 2018) devised an approach in which clustering is executed on input data to detect the structure of the data space. This is followed by a semi-supervised learning method to carry out classification using clustering information. Peikari and Martel (Peikari and Martel, 2016) propose a color transformation step that maps the Red-Green-Blue color space by computing eigenvectors of the RGB space. The color mapped image is then used in cell segmentation. Chen et al. (Chen et al. 2017) propose a deep learning network that implements a multi-task learning framework through multi-level convolutional networks for detection and segmentation of objects in tissue images.

Despite a large body of research work on image classification and segmentation, the process of extracting, mining, and interpreting information from digital slide images remains a difficult task (Xie et al. 2016a; Xing et al. 2016; Senaras, 2018; Chennubhotla *et al*., 2017). There are a number of challenges that segmentation and classification algorithms have to address. First, the morphology of tumor and normal tissue varies across tissue specimens – both across cancer types as well as across tissue specimens within a cancer type. Even a single tissue specimen will contain a variety of nuclei and other structures. Algorithms have to take into account tissue heterogeneity and learn and dynamically adapt to variations in tissue morphology across tissue specimens. It is not uncommon that an algorithm using fixed input parameters will do well for an image but poorly for another one. Second, nuclei in a tissue image touch or overlap each other. This is both a result of biological processes and an artifact of image capture. A tissue slide will have some depth, however small it is. Scanning a tissue specimen through a digitizing light microscope may inadvertently capture nuclei in different focal planes. Clumped nuclei make the segmentation process difficult. Third, whole slide tissue images are very high-resolution images and will not fit in main and GPU memory on most machines. Thus, it may not be feasible for a classification algorithm to work on an entire image as a whole. Algorithms have to be designed to work on at multiple resolutions or image tiles.

In this paper we present and experimentally evaluate two novel algorithms, one devised for segmentation of nuclei and the other developed for classification of whole slide tissue images:

- The segmentation algorithm proposes a multiscale deep residual aggregation network for accurate segmentation of nuclei and separation of clumped nuclei. Our method consists of three main steps. It first detects nuclear blobs and boundaries via a group of CNNs. It then applies a watershed algorithm on the results from the first step to perform an initial separation of clumped nuclei. The last step carries out a refined segmentation of separate nuclei from the second step. The proposed method employs a multi-scale approach in order to improve the detection and segmentation performance, because the sizes of nuclei vary across tissue specimens and within a tissue specimen. An evaluation of the segmentation algorithm using a set of image tiles from glioblastoma multiforme (GBM), lower grade glioma (LGG), head and neck squamous cell carcinoma (HNSCC) and non-small cell lung cancer (NSCLC) cases showed that the algorithm was able to achieve a segmentation accuracy of 0.78. The algorithm is not only able to accurately segment nuclear material but also separate touching and overlapped nuclei into individual objects.

- The classification algorithm proposes a two-part automated method to address the challenge of classifying non-small cell lung cancer (NSCLC) histology images. This method first classifies all input patches from an unseen WSI as NSCLC adeno (LUAD) or NSCLC squamous cell (LUSC) or non-diagnostic (ND) and obtains the corresponding probability maps for each class. Next, it extracts a collection of statistical and morphological features from the LUAD and LUSC probability maps as input into a random forest regression model to classify each WSI. This method is the first 3-class network that aims to classify each WSI into diagnostic and non-diagnostic areas. The experimental results with a set of

These two algorithms achieved the highest scores in the Computational Precision Medicine digital pathology challenge organized at the 20th International Conference on Medical Image Computing and Computer Assisted Intervention 2017 (MICCAI 2017), Quebec City, Canada. This challenge was organized by some of the co-authors of this manuscript to provide a platform for evaluation of classification and segmentation algorithms and is part of a series of annual digital pathology challenges organized since 2014. The 2017 challenge targeted tissue images obtained from patients with non-small cell lung cancer (NSCLC), head and neck squamous cell carcinoma (HNSCC), glioblastoma multiforme (GBM), and lower grade glioma (LGG) tumors. These cancer types are complex and deadly diseases accounting for a large number of diagnostic patient deaths in spite of application of various treatment strategies. In addition to methodology contributions presented in this paper, we will make the datasets used in the MICCAI 2017 Digital Pathology challenge publicly available for other researchers to use.

The rest of the manuscript is organized as follows. In Section 2 we introduce the nucleus segmentation algorithm and the classification algorithm. We present the experimental evaluation of the algorithms in Section 3. We describe the MICCAI 2017 Digital Pathology challenge and the challenge datasets in the same section. We conclude in Section 4.

## 2. MATERIALS AND METHODS

### 2.1 Segmentation of nuclei by a deep learning method

We developed an approach of convolutional neural networks (CNNs) to precisely segment nuclei. The method is composed of three major steps: 1) nuclei blob and boundary detection via CNNs, 2) separation of touching (or overlapping) nuclei by combining the nuclei blob and boundary detection results through a watershed algorithm, and 3) final segmentation of individual nuclei. The entire workflow is shown in Figure 1.

Two CNNs are trained to perform the initial nuclei blob and boundary detection. These CNNs consist of two consecutive processing paths – contracting path and expanding path – and are aimed at obtaining all the nuclei pixels and nuclei boundary pixels within the tissue image, generating nuclei blob and border masks, respectively. Provided with the blob and border masks, the initial nuclei segmentation is performed in two stages: 1) removal of the identified nuclei boundaries and 2) separation of the remaining clumped nuclei. In order to remove nuclei boundaries, we simply subtract the border mask from the blob mask after dilating the border mask with a kernel of size 3x3. A watershed algorithm is then applied to identify individual nuclei cores. Subsequently, each of the removed boundary pixels are assigned to its closest nuclei core, resulting in the segmentation of individual nuclei. Meanwhile, the size of each nuclei blob is examined to eliminate artifacts ($>13\mu m^2$).

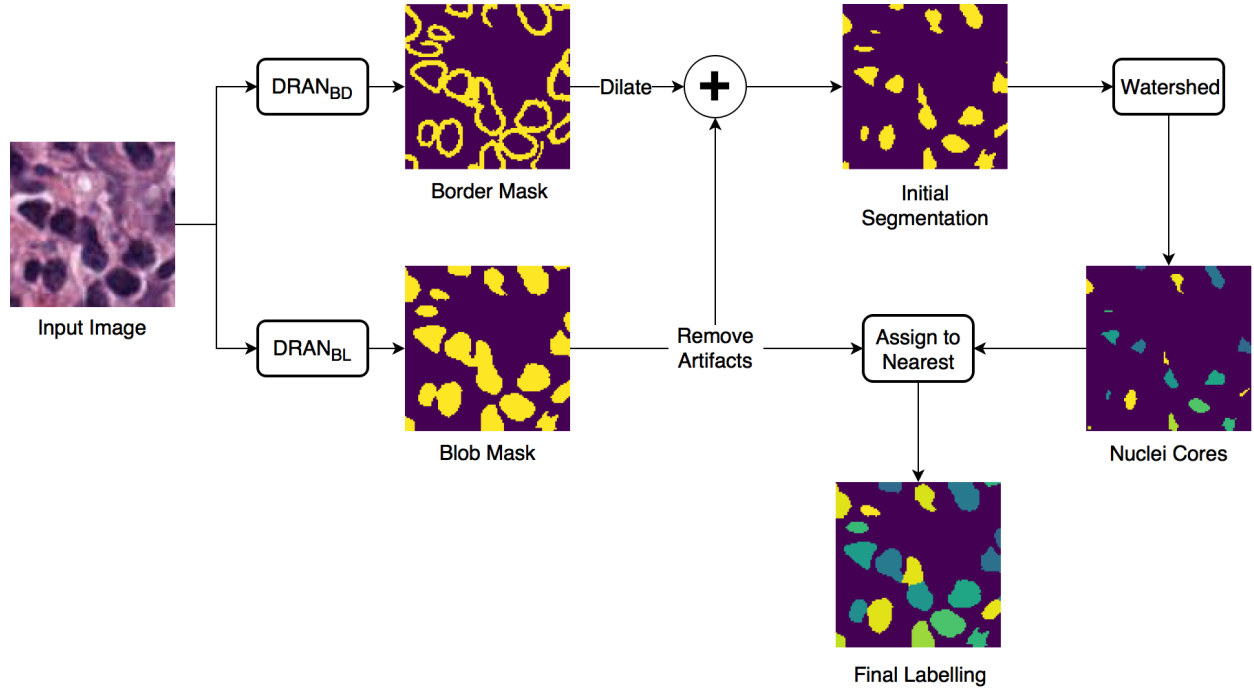

**FIGURE 1:** Overview of the nuclei segmentation procedure. $DRAN_{BL}$ and $DRAN_{BD}$ are the models for nuclei blob detection and boundary detection, respectively.

### 2.1.1 Network architecture

The proposed deep residual aggregation network (DRAN) is illustrated in Figure 2. It follows the renowned paradigm of two consecutive processing paths: contracting path (down-sampling the input) and expanding path (up-sampling the output of contracting path), such as in U-Net (Ronneberger *et al*., 2015), SegNet (Badrinarayanan *et al*., 2017), FCN (Long *et al*., 2015), and Hypercolumns (Hariharan *et al*., 2015), with several major and minor modifications.

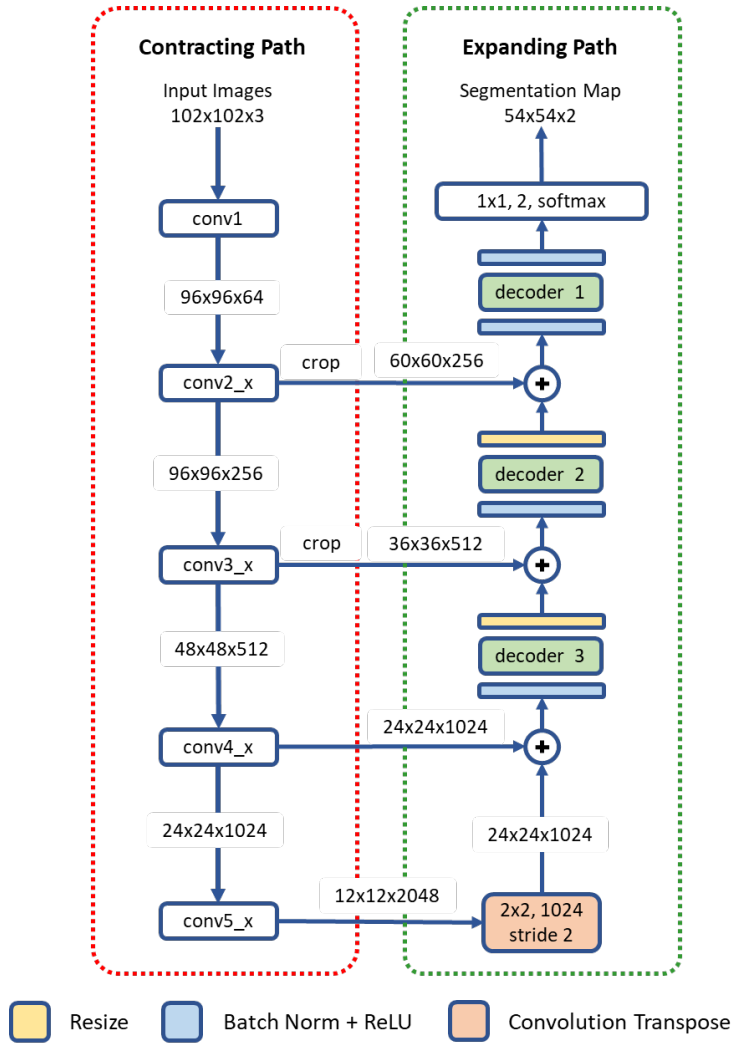

**FIGURE 2:** Architecture of a deep residual aggregation network (DRAN) for nuclei segmentation. Modified pre-activated ResNet50 is used for the contracting path.

### 2.1.2 Contracting path

The contracting path can be seen as a feature extraction step, recognizing the approximate position of the targeted objects and encoding their local characteristics. For this purpose, we utilize pre-activated ResNet50 (He *et al.*, 2016a). Unlike the original ResNet50 (He *et al.*, 2016) which uses the layout of convolution – batch normalization – rectified linear unit (ReLU) for residual units, the pre-activation architecture instead adopts the layout of batch normalization – ReLU – convolution and facilitates the direct propagation of the input via the shortcut path. Several modifications are made to the pre-activated ResNet50; the first 7x7 convolution is performed with a stride 1 and no-padding. The max pooling, following the first 7x7 convolution, has been removed.

### 2.1.3 Expanding path

The expanding path comprises four processing (or decoding) layers. The first layer receives the output of the last layer of the contracting path and performs transpose convolution and up-sampling. The second, third, and fourth layers receive two inputs – one from the preceding layer and the other from the contracting path. The two inputs are added together and go through a decoder and resizing unit. Unlike U-Net or DCAN (Chen *et al.*, 2017), the resizing unit simply doubles the size of the input with the nearest neighbor interpolation, which is computationally inexpensive. Moreover, instead of using concatenation as in (Ronneberger *et al.*, 2015), adopting addition operators reduces memory usage without substantially losing the learning capability of the network. The decoder is a primary processing unit that plays a key role in interpreting information from differing levels of abstraction and producing finer segmentation maps in the expanding path. It performs a series of convolution operations by employing multipath architecture (Xie *et al.*, 2017), where the input and output channels are divided into a number of disjoint groups (or paths), and each separately performs convolution. All the convolution operations in the decoder use no padding and a stride 1. Due to the convolution with no padding, the size of the segmentation map becomes smaller than that of an input image. Three decoders are utilized in our network. They share the same layout, but with differing number of channels and paths. The details are provided in Table 1.

**TABLE 1.** Details of three decoders. *C* is the number of paths (or groups). [5x5, 1024] denotes a kernel size of 5x5 and 1024 channels.

| decoder3 | decoder2 | decoder1 |
|---|---|---|
| $\begin{bmatrix} 5x5, & 1024, & \\ 3x3, & 1024, & C=256 \\ 1x1, & 512, & \end{bmatrix}$ | $\begin{bmatrix} 5x5, & 512, & \\ 3x3, & 512, & C=128 \\ 1x1, & 256, & \end{bmatrix}$ | $\begin{bmatrix} 5x5, & 256, & \\ 3x3, & 256, & C=64 \\ 1x1, & 128, & \end{bmatrix}$ |

### 2.1.4 Multiscale aggregation

The sizes of nuclei substantially vary among tissue samples, even within a single specimen. To better characterize nuclei and improve segmentation performance, we adopt a multiscale approach. The tissue specimen images are resized by a factor of 2 and 0.5. The resized images are separately fed into DRANs. Hence, three DRANs are, in total, trained and prepared: one at the original scale (x1.0) and the other two are at x2 and x0.5 scales. The last softmax layer of each DRAN is removed, and the output of decoder1 is aggregated through another decoder (*decoder4*), generating the final segmentation map at the original scale. We note that *decoder4* uses padding convolution. The details of the multiscale architecture are illustrated in Figure 3. This network is called as multiscale deep residual aggregation network (MDRAN).

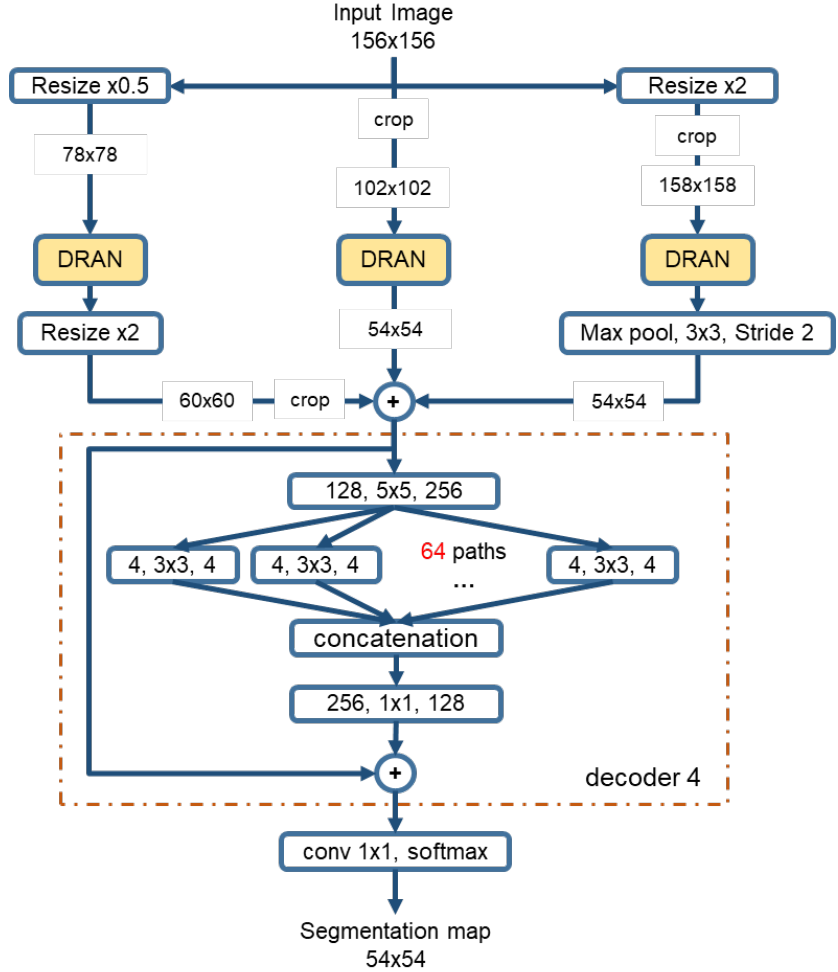

**FIGURE 3.** Multiscale Deep Residual Network (MDRAN) architecture. MDRAN composes of 3 DRANs at 3 scales (x0.5, x1.0, x2.0) and a decoder (in dash rectangle), aggregating 3 scales together and generating a segmentation map at x1.0 scale. In the decoder, the convolution block [128, 5x5, 256] denotes [128 input channels, 5x5 kernel, 256 output channels].

## 2.2 Classification of whole slide images by a two-part automated method

In recent years, there have been a number of published methods for automated NSCLC classification. Yu *et al*. (Yu et al. 2016) extracted a range of quantitative image features from tissue regions and used an array of classical machine techniques to classify each WSI. Although hand crafted approaches perform well, there is a growing trend towards deep learning approaches, where networks are capable of learning a strong feature representation. As a result of this strong feature representation, recent deep networks (He *et al*., 2016a; Szegedy *et al*., 2015; Simonyan *et al*., 2014; Huang *et al*., 2017) have achieved remarkable accuracy in large-scale image recognition tasks (Deng *et al*., 2009). Most WSI classification methods use a patch-based approach due to the computational difficulty in processing multi-gigapixel images. Coudray *et al*. (Coudray *et al*., 2018) classified NSCLC WSIs using deep learning on a patch-by-patch basis, but also predicted the ten most commonly mutated genes. For lung cancer classification, the authors used an Inception v3 network architecture to classify input patches into LUAD, LUSC and normal. They assumed that all patches within each WSI had the same label and therefore did not differentiate between diagnostic and non-diagnostic regions. This method may result in a large number of false positives in non-diagnostic regions and training may take a long time to converge. Hou *et al*., (Hou et al. 2016) trained a patch-level classifier to classify glioma and NSCLC WSIs into different cancer types. This was done by aggregating discriminative patch-level predictions from a deep network using either a multi-class logistic regression model or support vector machine. The selection of discriminative patches was done in a weakly supervised manner, where an expectation- maximization approach was used to iteratively select patches. These patches were then fed into a conventional two-class CNN to classify input patches as LUAD or LUSC. The authors of this method counter the problem of differentiating diagnostic and non-diagnostic regions by only considering discriminative patches. Although successful, this technique would likely fail if presented with a small unrepresentative dataset.

As a result of the above shortcomings, we present a method for non-small cell lung cancer classification, that primarily focuses on the diagnostic areas within the image for determining the cancer type. In section 2.2.1, we describe the deep learning framework for patch-based classification. In section 2.2.2 and 2.2.3 we describe the random forest regression model for classifying a whole slide image as LUAD or LUSC. A high-level overview of the classification framework can be viewed in Figure 4.

### 2.2.1 Network architecture

Inspired by the success of ResNet (He *et al*., 2016) in image-recognition tasks (Huang *et al*., 2017), we implemented a deep neural network with residual blocks at its core to classify NSCLC input patches. This network architecture is a variant of ResNet50, as described by He *et al*. (He *et al.* 2016), but we use a 3×3 kernel as opposed to a 7×7 kernel during the first convolution and reduce the number of parameters throughout the network. Using a 3×3 kernel is important in this domain because a smaller receptive field is needed to locate small features that are common in histology images. Reducing the number of parameters allows the network to be more generalized and reduces the possibility of over-fitting. In order to reduce the number of parameters, we modified ResNet50 (He *et al*., 2016) by reducing the number of residual blocks throughout the network so that we had 32 layers as opposed to 50. Due to the high variability between images, and therefore between the training and validation set, consideration for preventing over-fitting is crucial. Figure 5 gives an overview of the network architecture.

Once training was complete, we selected the optimal epoch corresponding to the greatest average validation accuracy and processed patches from each test WSI. This resulted in three probability maps; one for each class.

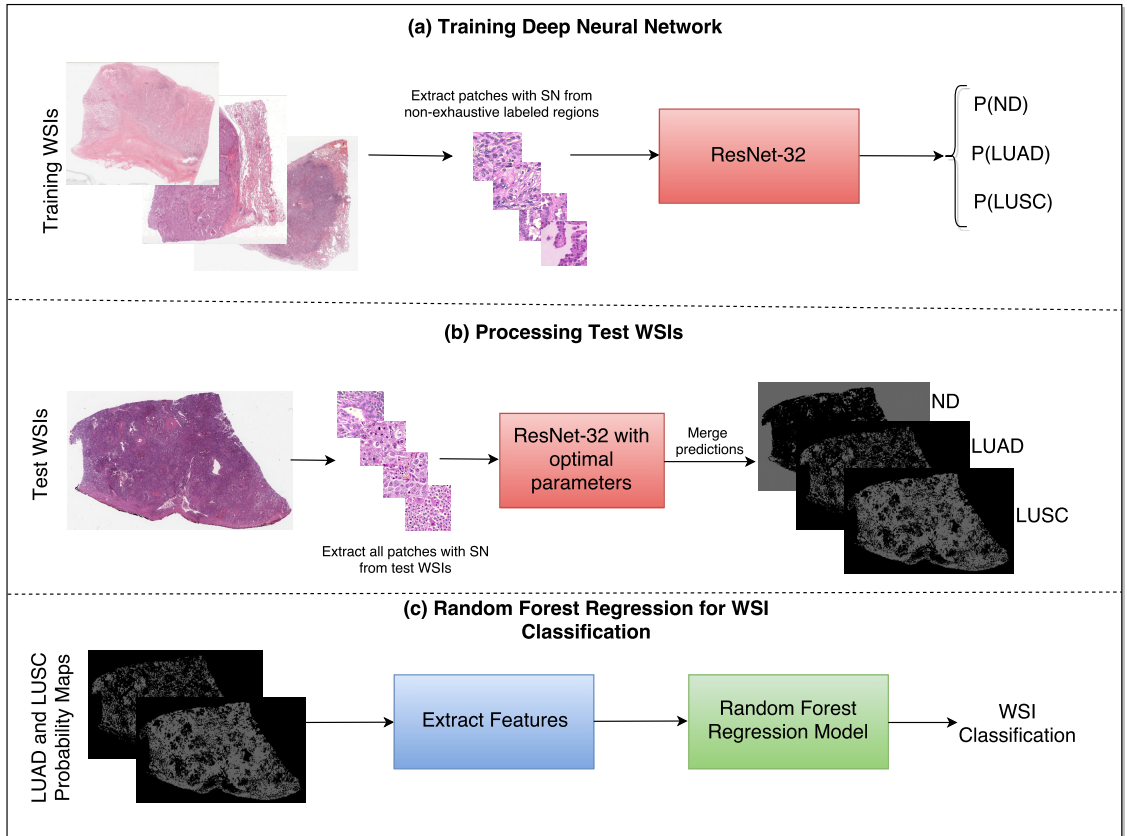

**FIGURE 4:** Overview of the NSCLC classification framework. (a) Workflow for training the neural network to classify input patches as either non-diagnostic (ND), lung adenocarcinoma (LUAD) or lung squamous cell carcinoma (LUSC). (b) Workflow for processing the WSIs within the test set to obtain probability maps for each class. (c) Workflow for the random forest regression model. Features are extracted from LUAD and LUSC probability maps and then fed as input into the random forest model. SN stands for stain normalization by method of Reinhard (Reinhard *et al*., 2001).

### 2.2.2 Extraction of statistical and morphological features

For classifying each WSI as either lung adenocarcinoma or lung squamous cell carcinoma, we extracted features from both the LUAD and LUSC probability maps. We explored two post processing techniques: max voting and a random forest regression model. Max voting simply assigns the class of the WSI to be class with the largest number of positive patches in its corresponding probability map. Therefore, max voting only requires the positive patch count for both the LUAD and LUSC probability maps in order to make a classification. For the random forest regression model, we extracted 50 statistical and morphological features from both the LUAD and LUSC training probability maps and then selected the top 25 features based on class separability. We gained the training probability maps by processing each training WSI with a late epoch. This ensured that the network had over-fit to the training data and gave a good segmentation of LUAD and LUSC diagnostic regions. In other words, using this method allowed us to transition from a non-exhaustive to an exhaustive labeled probability map. Once the model was trained with these features, they were then input as features into the random forest regression model. Statistical features that were extracted included: mean, median and variance of the probability maps. We also calculated the ratio between the LUAD and LUSC probability maps. Morphological features that were extracted included the size of the top five connected components at different thresholds.

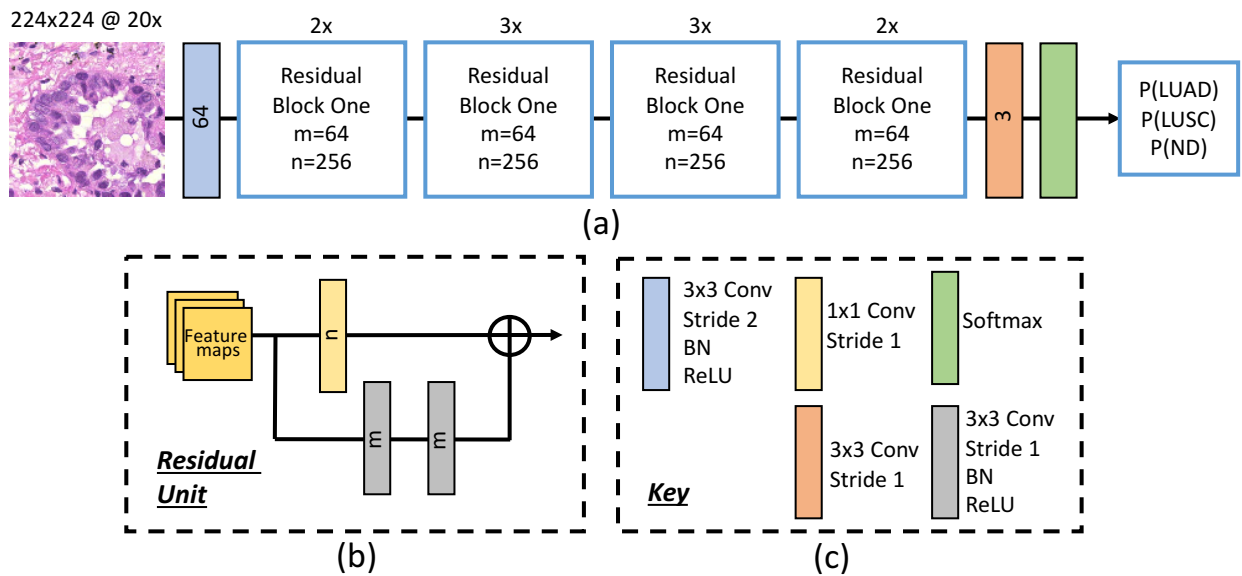

**FIGURE 5:** The deep convolutional neural network. (a) Network architecture, (b) residual unit. Within the residual block, ⊕ refers to the summation operator. (c) Key highlighting each component within the workflow. Note, the number within each convolutional operator denotes the output depth. Above each residual block we denote how many residual units are used.

### 2.2.3 Random forest regression model

An ensemble method is a collection of classifiers that are combined together to give improved results. An example of such an ensemble method is a random forest, where multiple decision trees are combined to yield a greater classification accuracy. Decision trees continuously split the input data, according to a certain parameter until a criterion is met. Specifically, a random forest regression model fits a number of decision trees on various sub- samples of the data and then calculates the mean output of all decision trees. We optimized our random forest model by selecting an ensemble of 10 bagged trees, randomly selecting one third of variables for each decision split and setting the minimum leaf size as 5. We finally selected a threshold value to convert the output of the random forest regression model into a binary value, indicating whether the WSI was LUAD or LUSC.

## 3. RESULTS

### 3.1 Digital pathology challenge and datasets

We organized the MICCAI 2017 digital pathology challenge to provide a venue for comparing algorithms using a common, curated set of datasets and help in advancing algorithm development in digital pathology. The 2017 challenge consisted of two sub-challenges; segmentation of nuclei in tissue images and the classification of whole slide tissue images (WSIs). It used tissue images obtained from patients with non-small cell lung cancer (NSCLC), head and neck squamous cell carcinoma (HNSCC), glioblastoma multiforme (GBM), and lower grade glioma (LGG) tumors. These cancer types are complex and deadly diseases accounting for a large number of diagnostic patient deaths in spite of application of various treatment strategies.

### 3.1.1 Segmentation of nuclei in images

In this sub-challenge, challenge participants were asked to apply automated algorithms to detect and segment all of the nuclei in a set of tissue images. The tissue image dataset consisted of image tiles extracted from whole slide tissue images. Image tiles were used instead of whole slide tissue images because of the significant time and resource cost of manually and accurately segmented nuclei a WSI. In our experience, a WSI may have hundreds of thousands to millions of nuclei. It would be infeasible to generate a ground truth dataset from even a single WSI, let alone from tens of WSIs. In addition, processing a WSI for nucleus segmentation may require significant computing power. Using tiles instead of WSIs in the challenge reduced computational and memory requirements, as the primary objective of the challenge was to evaluate the accuracy performance of an algorithm.

This sub-challenge used images from The Cancer Genome Atlas (TCGA) repository (TCGA, 2018). The image tiles for the training and test sets were selected from a set of GBM, LGG, HNSCC, and NSCLC whole slide tissue images by Pathologists and extracted using Aperio's ImageScope software. The training set and the test set each consisted of 32 image tiles with 8 tiles from each cancer type. We recruited a group of students to manually segment all the nuclei in the image tiles. Each tile was segmented by multiple students using a desktop software called iPhotoDraw (http://iphotodraw.com). The student segmentations were reviewed by Pathologists in review sessions with the students. In the review sessions, the manual segmentations were refined, and a consensus segmentation was generated for each image tile. Then labeled masks were generated to represent manual segmentations. A labeled mask represents each segmented nucleus in an image tile with a different id. All the pixels that are part of the same nucleus are assigned the same id.

The score of a segmentation output was computed using the DICE coefficient (DICE_1) (Dice, 1945) and a variant of the DICE coefficient which we implemented and called "Ensemble Dice" (DICE_2).  The DICE coefficient measures overlap between ground truth and algorithm segmentation output but does not take into splits and merges. A "Split" is the case in which the human segments a region in a single nucleus, but the algorithm segments the same region in multiple nuclei. A "Merge" is the case in which the algorithm segments a region in a single nucleus, but the human segments the same region in multiple nuclei.  With the DICE coefficient, an algorithm that segments two touching (or overlapping) nuclei as a single object will have the same DICE_1 value as an algorithm that correctly segments the nuclei as two separate objects.

DICE_2 was implemented to capture mismatch in the way ground truth and an algorithm segmentation of an image region are split. The pseudo-code for DICE_2 is given below.

---
**Algorithm.** Computing Ensemble Dice Coefficient (DICE_2)
---
**Input:** An image contains a set of segmented nuclear Q where each nuclei is indexed by $q$ and a second image contains a sets of segmented nuclear P where each nuclei is indexed by $p$.
**Output:** Ensemble Dice Score *DICE_2*
1: Initialize total intersection and markup pixels count:
   IntersectionArea ← 0;
   TotalMarkupArea ← 0;
2: **foreach** $q$ **in** $Q$ **do**:
3:    **foreach** $p$ **in** $P$ **do**:
4:       **if** $q$ intersects $p$ **then**:
            IntersectionArea ← IntersectionArea + AreaOfOverlap(q,p)
            TotalMarkupArea ← TotalMarkupArea + (Area(q) + Area(p))
5:       **end if**
6:    **end for**
7: **end for**
8: *DICE_2* ← 2 * IntersectionArea / TotalMarkupArea

---

Here, Q and P are the sets of segmented objects (nuclei). The two DICE coefficients were computed for each image tile in the test dataset. The score for the image tile was calculated as the average of the two dice coefficients. The score for the entire test dataset was computed as the average of the scores of all the image tiles.

### 3.1.2 Classification of whole slide tissue images

This sub-challenge used images from NSCLC cases. All the images were also obtained from whole slide tissue images in TCGA repository Each whole slide tissue image stored in the TCGA repository has diagnostic information about the category of cancer tumor (e.g., gbm, lgg, ovarian) as well as associated clinical outcome data and genomics data. The images were reviewed and selected by a pathologist. In the NSCLC cases, the images were selected from NSCLC adeno (LUAD) and NSCLC squamous cell (LUSC) cases. Each case in the dataset had one image – so a classification of images would correspond to a classification of cases. Challenge participants were asked to apply their algorithms to classify each image as NSCLC adeno or NSCLS squamous cell. The training dataset had a total of 32 cases; 16 LUAD and 16 LUSC cases. The test dataset had a total of 32 cases with 16 LUAD and 16 LUSC cases. The images were made available in the original file format (i.e., Aperio svs format). The original TCGA filename of each image was mapped to a generic filename (i.e., image1.svs, image2.svs, etc). The label image and image metadata showing the TCGA case id were removed from the image files. Ground truth was supplied for the training images that gave the cancer type of each WSI, whereas this ground truth was held back for the test images. The score of an analysis algorithm was computed as the number of correctly classified cases divided by the total number of cases.

## 3.2 Experimental evaluation of deep learning method for segmentation of nuclei

From the original 32 training image tiles, with no additional preprocessing steps, multiple patches (~100 per image) of size 200x200 are extracted. Three training datasets are generated (Table 2). By sliding a window with a step of 54 pixels and random cropping, 4732 patches are generated, designated as Nuclei Blob (NBL) dataset and used for nuclei blob detection. Nuclei Boundary (NBD) dataset is generated by centering each nucleus at the center of each patch, producing 2785 patches that are used for nuclei boundary detection. Small Nuclei (SN) dataset is the duplicate dataset of NBL that only contains nuclei blob patches possessing ≤ 50% nuclei pixels. SN dataset is only used for training DRAN for nuclei blob detection.

During training, data augmentation is applied as follows: 1) a random vertical and horizontal shift in a range of [-0.05, 0.05] with respected to the patch's width and height 2) a random rotation in a range of [-45°, 45°] degree 3) a random vertical and horizontal flipping with probability 0.5 4) a random shear with intensity in a range of [-0.4π, 0.4π] 5) a random resizing with a ratio in a range of [0.6, 2.0]. This augmentation is to address variations of nuclei in contrast, shapes, and etc. that are often observed in pathology images. This is known to be helpful in coping with the natural variations present in the images as well as ensuring the robustness of the network (Ronneberger *et al*., 2015). Following the augmentation, the center region of size 102x102 is extracted prior to being fed into the network (Figure 6). Augmentation is performed 3 times per patch.

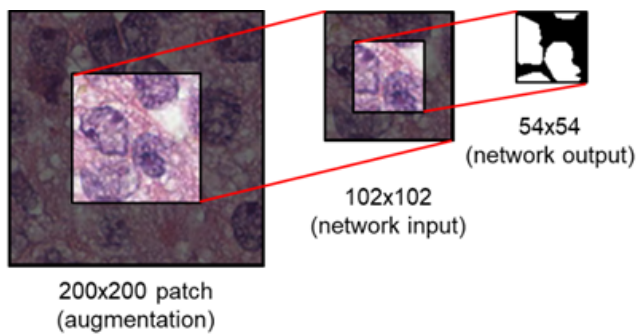

**FIGURE 6:** Image patch generation. To avoid zero-padding in augmentation, a patch of size 200x200 is first provided. Subsequently, the center region of 102x102 is cropped and fed into the network as input. For an input of size 102x102, the network provides a segmentation map of size 54x54.

Using the generated training data above, DRAN is trained via Adam optimizer with default parameter values ($\beta_1$=0.9, $\beta_2$=0.999, $\epsilon$=1e-8). A mini batch size of 32 is maintained throughout the whole training process. L2 regularization loss is also applied with a factor of 1.0e-5 to improve the generalizability of the proposed network. *K.He initialization* (He *et al*., 2015a) is utilized to initialize the weights for the convolutional layers in the expanding path. Training is performed in two phases. In the first phase (35 epochs), the pretrained weights of pre-activated ResNet50 are loaded into the contracting path and is kept frozen (no update on weights), i.e., only the expanding path is trainable in this phase. The learning rate is initially set to 1.0e-4, then changes to 5.0e-5, 1.0e-5, 7.5e-6 and 5.0e-6 at the $0^{th}$, $1^{st}$, $15^{th}$, $25^{th}$ and $35^{th}$ epoch, respectively. In this phase, the network is trained with NBL dataset for nuclei blob detection and NBD dataset for nuclei boundary detection. In the second phase (40 epochs), the contracting path is unfrozen, that is, the whole network becomes trainable. For nuclei blob detection, both NBL and SN datasets are used to further refine the network. Only NBD dataset is utilized for nuclei boundary detection. In addition, differing penalties for the loss function are imposed to alleviate the heavy bias in NBD dataset; each border pixel has a weight of 5.0 and miss-classifying a border pixel as background gains a weight 6.0. Background pixels have 1.0 weight while miss-classifying them is penalized with 4.0.

**TABLE 2.** Generation of three datasets from the original 32 image tiles.

| Dataset | Number of Patches | Extraction Details |
|---|---|---|
| Nuclei Blob (NBL) | 4732 | • Window-slide cropping with a step of 54 pixels per image tile<br>• Random-cropping per image tiles, 30 times |
| Nuclei Boundary (NBD) | 2785 | • Each patch centering a single nucleus<br>• Nuclei near image tile's edges are ignored |
| Small Nuclei (SN) | 14552 | • Duplicate patches from NBL which contain ≤ 50% nuclei pixels in the center region of 54x54, 3 times<br>• Only for training nuclei blob detection |

On the other hand, with the same training data as DRAN, the training procedure of the multiscale model is detailed as followed: each DRAN branch of the MDRAN is loaded with the pretrained DRAN weights that are obtained from the procedure described above and is kept frozen. The network then proceeds to train *decoder4* for 10 epochs with the learning rate of 1.0e-4. Afterwards, the expanding path of DRANs is unfrozen and finetuned for additional 35 epochs while the learning rate is set to be 1.0e-4, 1.0e-5 and 1.0e-6 at the $1^{st}$ epoch, $15^{th}$ epoch and $30^{th}$ epoch, respectively.

**TABLE 3.** Segmentation of nuclei performance.

| Method | DICE_1 | DICE_2 | Average Score |
|---|---|---|---|
| $DRAN_{BL}+DRAN_{BD}$ | 0.8532 | 0.7010 | 0.777 |
| $MDRAN_{BL}+DRAN_{BD}$ | **0.8620** | **0.7033** | **0.783** |

Overall, utilizing NBL+SN dataset, $MDRAN_{BL}$ was trained for nuclei blob detection. $DRAN_{BD}$ was trained on NBD dataset for nuclei boundary detection. Combining the two models ($MDRAN_{BL}+DRAN_{BD}$), nuclei segmentation was performed. Table 3 shows the segmentation results on the test set. Our method ($MDRAN_{BL}+DRAN_{BD}$) achieved 0.862 DICE_1, 0.703 DICE_2, and the average score of 0.783. The effect of the multiscale aggregation ($MDRAN_{BL}$) was examined. Using a single scale nuclei segmentation method ($DRAN_{BL}+DRAN_{BD}$), we obtained 0.853 DICE_1, 0.701 DICE_2, and the average score of 0.777, worse than those of the multiscale aggregation. In a head-to-head comparison of the test set, the multiscale aggregation substantially improved the segmentation performance, especially on three test images that were scanned at 20x magnification (Figure 7). As for other test images, scanned at 40x magnification, the multiscale aggregation, in general, slightly outperformed the single scale method. This suggests that the multiscale aggregation, in particular, aids in improving the segmentation of (relatively) smaller nuclei. Figure 8 shows the segmentation results by the multiscale aggregation and single scale method; the single scale method missed several small nuclei that were, however, identified by the multiscale aggregation.

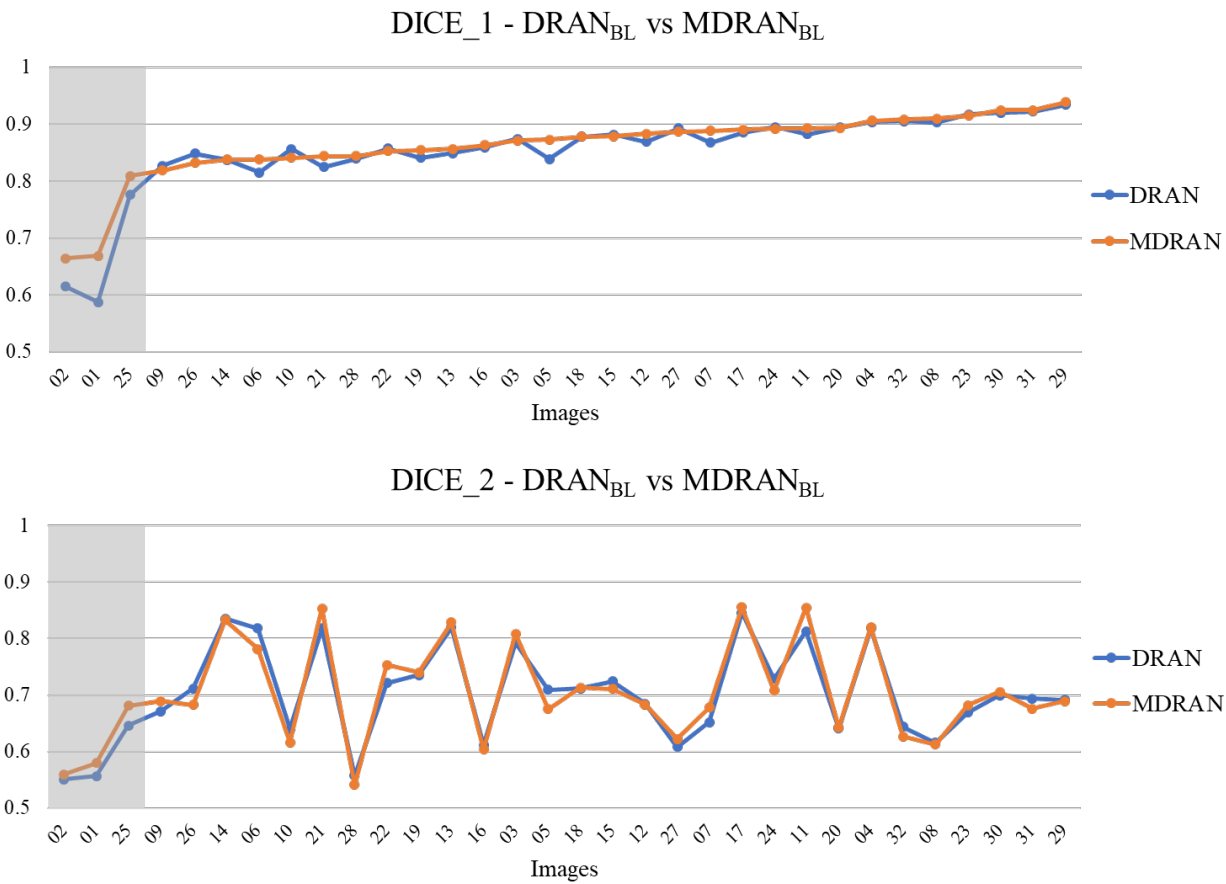

**FIGURE 7:** Head-to-head comparison between MDRAN$_{BL}$ and DRAN$_{BL}$ on the test set. Test images are ordered by the ascending order of MDRAN DICE_1. The shaded area indicates that the images were scanned at 20x magnification.

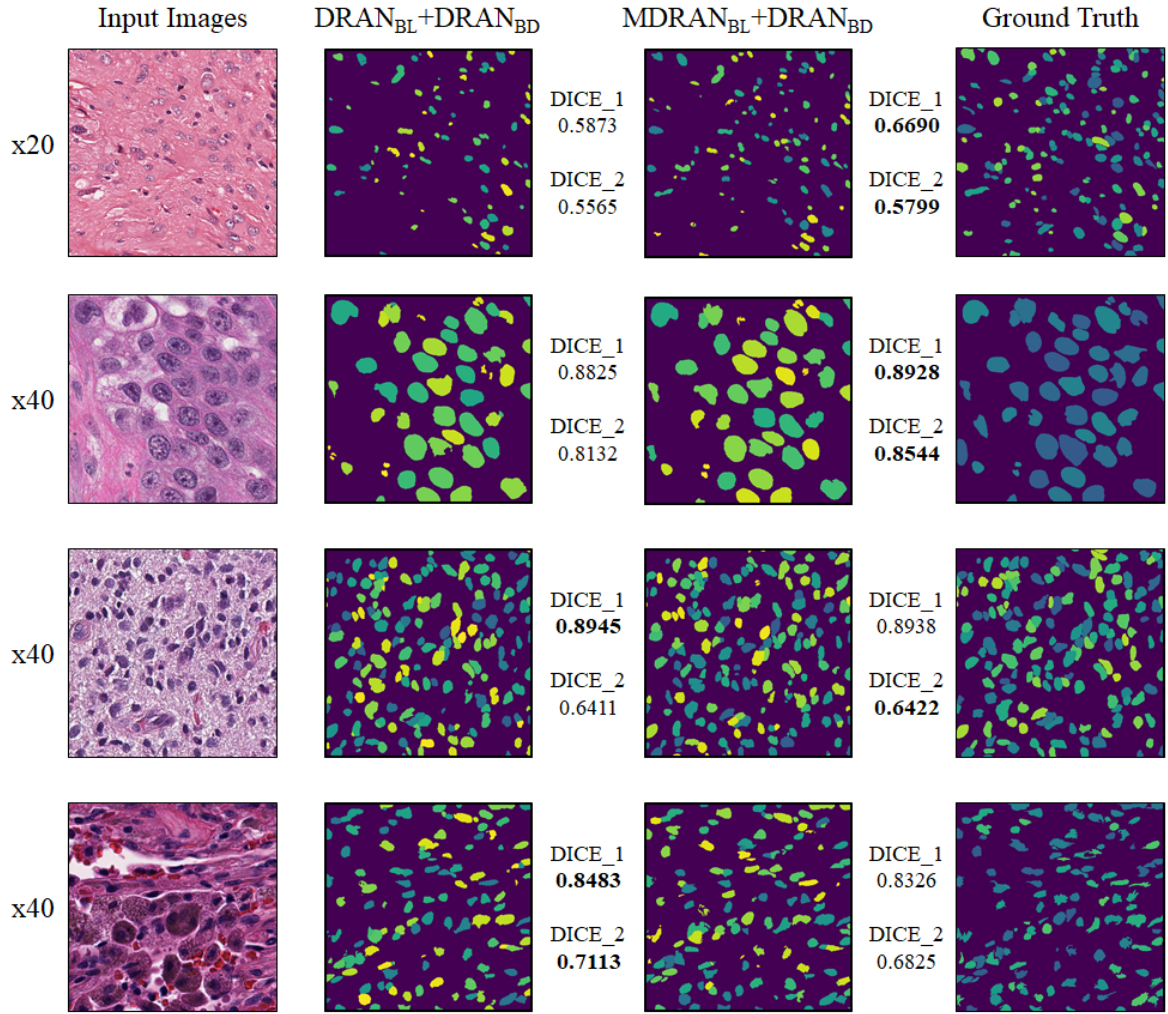

**FIGURE 8:** Examples of nuclei segmentation via the multiscale aggregation (MDRAN$_{BL}$+DRAN$_{BD}$) and single scale (DRAN$_{BL}$+DRAN$_{BD}$) approach. The images from top to bottom are the 1$^{st}$, 11$^{th}$, 20$^{th}$ and 25$^{th}$ image tile in the test set.

Notably, a huge discrepancy between DICE_1 and DICE_2 was observed for several test images (Figure 7). Upon closer inspection, we found that these are mainly due to staining variation and instability as well as densely overlapping nuclei. As shown in Figure 9, the identified nuclei boundaries are often fragmented and imperfect, leading to inaccurate segmentation of the overlapping nuclei. This indicates that advanced and sophisticated touching nuclei separation method may hold a great potential for improving the segmentation performance.

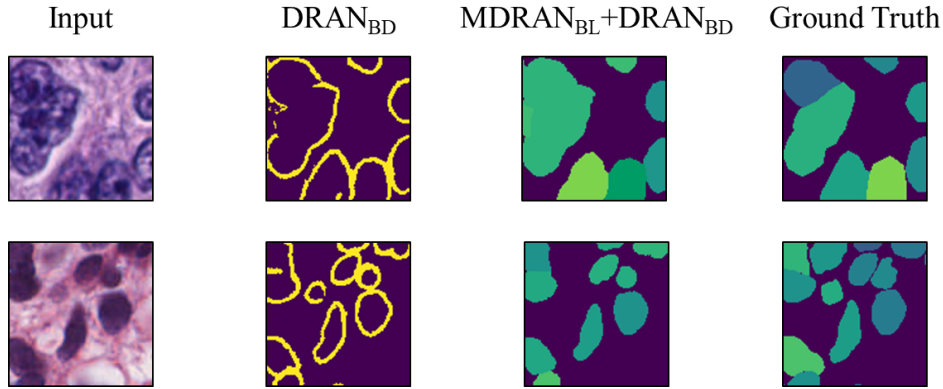

**FIGURE 9:** Examples of correct and incorrect nuclei segmentation. Our method (Bottom) is able to distinguish the boundary of the non-highly overlapping nuclei fairly well but (Top) fails on the highly overlapping nuclei with disproportionate stains.

### 3.3 Classification of Whole Slide Images

We used a total of 64 Hematoxylin and Eosin (H&E) NSCLC WSIs that were split into 32 training and 32 test images. We had an even breakdown of NSCLC images in both the training and the test set, giving a total of 32 LUAD slides and 32 LUSC slides. We divided our dataset so that we had 24 WSIs for training and 8 for validation, with 4 validation images taken from LUAD and LUSC respectively. We extracted a 3-class dataset comprising of patches of size 256×256 at 20× magnification, from non-exhaustive labeled regions, confirmed by an expert pathologist (AK). This 3-class dataset consisted of LUAD, LUSC and non-diagnostic areas (ND). LUAD diagnostic regions within the slide consisted of: tumor; growth pattern structures and tumor stroma. LUSC diagnostic regions consisted of: tumor; keratin pearls and tumor stroma. Non-diagnostic regions included: fat; lymphocytes; blood vessels; alveoli; red blood cells; normal stroma; cartilage and necrosis. We considered necrosis to be non-diagnostic because, despite LUSC generally having more necrotic areas than LUAD, it is not indicative of lung squamous cell carcinoma on a patch-by-patch basis. In this case, it was particularly important to incorporate non-diagnostic areas because there were no normal cases within the dataset. Overall, our network is optimized on 65,788 training image patches.

There was a high level of stain variation between all images, due to images being acquired from different centers. To counter this stain variability, we applied Reinhard (Reinhard *et al*., 2001) stain normalization to all images by mapping each image to the statistics of a pre-defined target image. During training we performed random crop, flip and rotation data augmentation to make the network invariant to these transformations. After performing a random crop to all input patches, we were left with a patch size of 224×224.

An increase in the amount of labeled data coupled with a surge in computing power has allowed deep convolutional neural networks to achieve state-of-the-art performance in computer vision tasks. The hierarchical architecture of such networks allows them to have a strong representational power, where the complexity of learned features increases with the depth of the network. The proposed network $f$ is a composition of a sequence of $L$ functions of layers $(f_1, \ldots, f_L)$ that maps an input vector $\boldsymbol{x}$ to an output vector $\boldsymbol{y}$, i.e,

$$\boldsymbol{y} = f(\boldsymbol{x}; \boldsymbol{w_1}, \ldots, \boldsymbol{w_L})$$
$$= f_L(.; \boldsymbol{w_L}) \circ f_{L-1}(.; \boldsymbol{w_{L-1}}) \circ \ldots \circ f_2(.; \boldsymbol{w_2}) \circ f_1(.; \boldsymbol{w_1}) \quad (1)$$

where $\boldsymbol{w_L}$ is the weight and bias vector for the $L^{th}$ layer $f_L$. In practice, $f_L$ most commonly performs one of the following operations: a) convolution with a set of filters; b) spatial pooling; and c) non-linear activation.

Given a set of training data $\{(\boldsymbol{x}^{(i)}, \boldsymbol{y}^{(i)})\}$, where $i$ ranges from 1 to N. We can estimate the vectors $\boldsymbol{w_1} \ldots \boldsymbol{w_L}$ by solving:

$$\underset{\boldsymbol{w_1},\ldots,\boldsymbol{w_L}}{\mathrm{argmin}} \frac{1}{N} \sum_{i=1}^{N} l(f(\boldsymbol{x}^{(i)}; \boldsymbol{w_1}, \ldots, \boldsymbol{w_L}), \boldsymbol{y}^{(i)}) \quad (2)$$

where $l$ is the defined loss function. We perform numerical optimization of (2) conventionally via the back- propagation algorithm and stochastic gradient descent methods.

In addition to the above operations, residual networks (ResNets) (He *et al*. 2016) have recently been proposed that enable networks to be trained deeper and as a result, benefit from a greater accuracy. Current-state-of-the-art networks (He *et al*., 2016; Szegedy *et al*., 2015; Simonyan *et al*., 2014; Huang *et al*., 2017) indicate that network depth is of crucial importance, yet within conventional CNNs, accuracy gets saturated and then degrades rapidly as the depth becomes significantly large. The intuition behind a residual network is that it is easier to optimize the residual mapping than to optimize the original unreferenced mapping. Residual blocks are the core components of ResNet and consist of a feed-forward skip connection, that performs identity mapping, without adding any extra parameters. These connections propagate the gradient throughout the model, which in turn enables the network to be trained deeper, often achieving greater accuracy.

Table 4 summarizes the experiments we carried out for classification of input patches into LUAD, LUSC and ND. We choose to train the specified networks, due to their state-of-the-art performance in recent image recognition tasks (Deng *et al*. 2009). During training, all networks quickly over-fit to the training data. This was because of two reasons: (i) The networks architectures that were used have been optimized for large-scale computer vision tasks with millions of images and thousands of classes; (ii) We have a fairly limited training set size. Due to the size of our dataset, (ii) it is difficult to avoid over-fitting, given a sufficient number of model parameters. Therefore, we modify the network architecture to counter the problem of over-fitting by reducing the number of layers. We make a modification to the original implementation of ResNet by reducing the number of residual units, such that we only have a total of 32 layers within the model. Modification of ResNet50 to give ResNet32 helped alleviate the problem of over-fitting and gave the best patch-level performance. Despite only achieving 0.4% greater accuracy than InceptionV3, ResNet32 resulted in a significantly greater average LUAD and LUSC patch-level accuracy. The average LUAD and LUSC patch-level accuracy for InceptionV3 was 0.678, whereas the average accuracy for ResNet32 was 0.776. As a consequence of the superior patch-level performance, we chose to use ResNet32 for processing images in the test set. Figure 10 shows four test WSIs with their overlaid probability maps. Green regions show regions classified as LUSC, blue/purple regions show regions classified as LUAD and yellow/orange regions show regions classified as ND.

Table 5 shows the overall accuracy for NSCLC WSI classification, as processed by the challenge organizers. We observe that using the random forest regression model with statistical and morphological features from the labeled WSI increases the classification accuracy. Max voting is sufficient when either LUAD or LUSC is a dominant class within the labeled WSI, but when there is no obvious dominant class, the random forest regression model increases performance. This is because the features used as input to the random forest model are more informative than simply using a voting scheme and can therefore better differentiate between each cancer type.

**TABLE 4.** Patch-level accuracy. LUAD refers to lung adenocarcinoma, LUSC refers to lung squamous cell carcinoma, ND refers to non-diagnostic area of interest.

| Network | Resolution | LUAD | LUSC | ND | Average |
|---|---|---|---|---|---|
| VGG | 20x | 0.634 | 0.663 | 0.826 | 0.708 |
| InceptionV3 | 20x | 0.623 | 0.733 | 0.924 | 0.760 |
| ResNet50 | 20x | 0.601 | 0.597 | 0.889 | 0.695 |
| ResNet32 | 20x | 0.702 | 0.849 | 0.742 | 0.764 |

**TABLE 5.** Overall WSI classification accuracy. ResNet32-MV refers to classifying input patches using ResNet32, then using majority voting as a post processing classification technique. ResNet32-RF refers to classifying input patches using ResNet32 and then using a random forest regression model as a post processing technique for classification.

| Method | Accuracy |
|---|---|
| ResNet32-MV | 0.78 |
| ResNet32-RF | 0.81 |

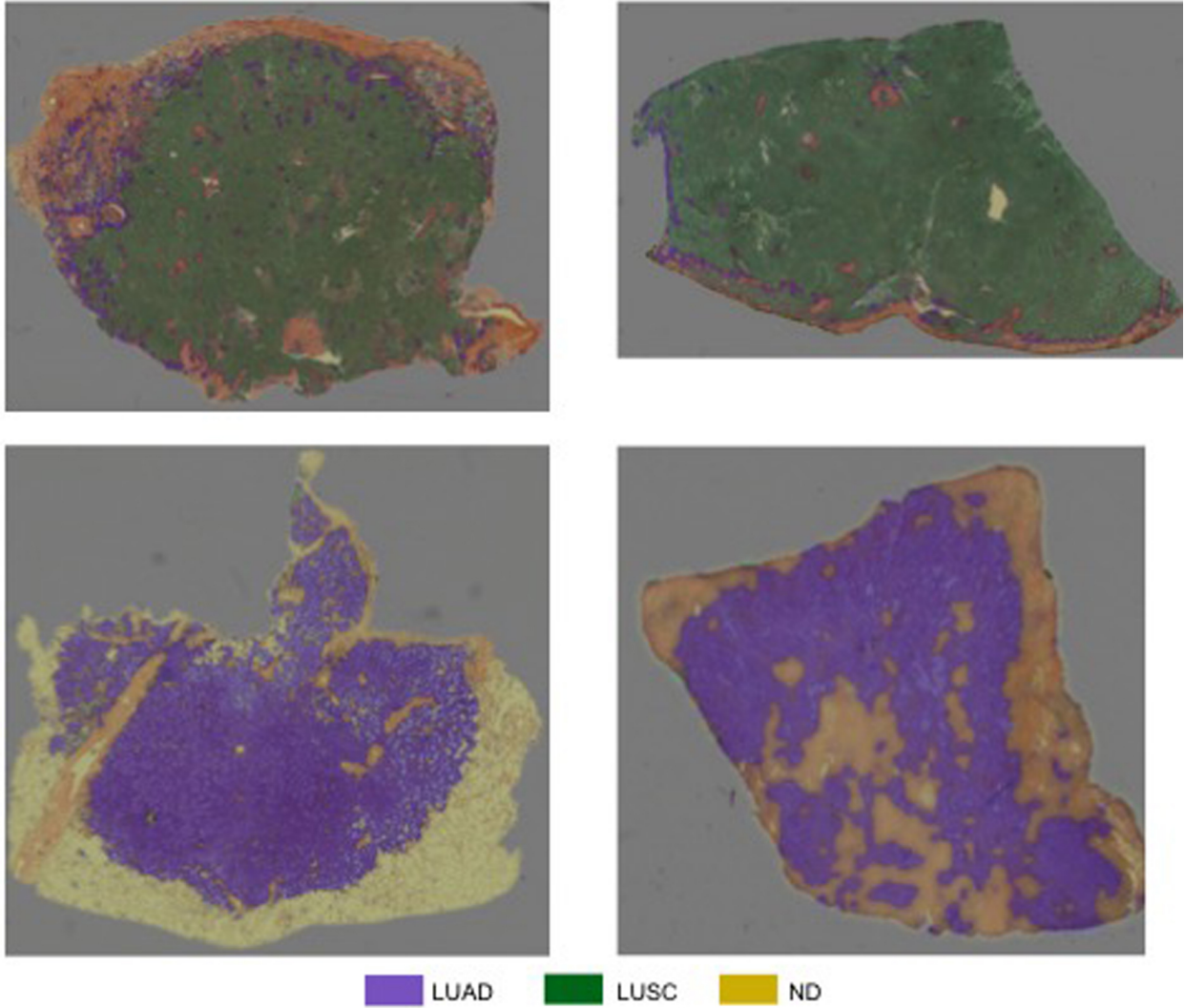

**FIGURE 10:** Test WSIs with overlaid probability maps. Blue/purple indicates a region classified as diagnostic LUAD, green indicates a region classified as diagnostic LUSC, yellow/orange refers to a region classified as non-diagnostic.

## 4. DISCUSSION

Understanding the interplay between morphology and molecular mechanisms is central to the success of research targeting practically every major disease. There are latent as well as observable changes in tissue structure and function at disease onset and over the course of disease progression. Traditionally whole slide tissues are manually examined under a high-power light microscope to render a diagnosis. This manual process is laborious, limiting the number of tissue samples that can be used in a study. Digital Pathology enables quantitative studies of these changes and underlying disease mechanisms at the sub-cellular scales. Integrating this information into the landscape of the entire spectrum of clinical information will drive both disease specific and patient specific information which can be used to drive high risk high reward cancer trials to better results faster. To bring this vision to reality it will be necessary to develop and deploy accurate and reliable analysis algorithms. The development of such algorithms remains to be a challenge. The primary challenge stems from the fact that tissue images contain much denser amount of data than many other imaging modalities and these data encode rich information at multiple scales. This challenge is compounded by the fact that most analysis methods are based on heuristics (e.g., to cope with data and computation requirements) and can be sensitive to input parameters. Advances in the accuracy, robustness, and efficiency of digital pathology image analysis methods and pipelines will be accelerated by systematic, critical evaluation of methods and through engagement of the community of method developers. In this paper we presented a compendium of two novel segmentation and classification methods contributed by the two top scoring participants in the "Computational Precision Medicine Digital Pathology Challenge" held at the MICCAI 2017 conference and the results from these methods.

The multiscale deep residual aggregation network consists of three main steps. It first detects nuclear blobs and boundaries via a group of CNNs. It then applies a watershed algorithm on the results from the first step to perform an initial separation of clumped nuclei. The last step carries out a refined segmentation of separate nuclei from the second step. The proposed method employs a multi-scale approach in order to improve the detection and segmentation performance, because the sizes of nuclei vary across tissue specimens and within a tissue specimen. The experimental evaluation of the method shows a combined DICE_1 and DICE_2 score of 0.78. The experimental evaluation suggests that (1) the multiscale aggregation aids in improving the segmentation of (relatively) smaller nuclei and (2) advanced and sophisticated touching nuclei separation methods may hold a great potential for improving the segmentation performance in tissue specimens with discernable staining variation and instability as well as densely overlapping nuclei.

The automated method for non-small cell lung cancer classification implements a deep neural network to classify input patches as lung adenocarcinoma, lung squamous cell carcinoma or non-diagnostic regions. Initially, WSIs are divided into image tiles and each image tile is classified into one of the three classes. Statistical and morphological features are subsequently extracted from the LUAD and LUSC probability maps and are then used as input into a random forest regression model to classify each WSI as lung adenocarcinoma or lung squamous cell carcinoma. The proposed method achieves the greatest accuracy with a score of 0.81 as part of the digital pathology challenge at MICCAI 2017, highlighting the superior performance of our classification framework. Given the limitation of the dataset, it is clear that classifying NSCLC WSIs into diagnostic and non-diagnostic regions is of crucial importance. This is particularly important for this specific implementation because the training set did not contain any normal cases. Without the consideration of non-diagnostic areas, the algorithm would be forced to make a prediction for non-informative image tiles. Furthermore, it is evident that the analysis of the morphology of classified regions can empower the classification of the whole-slide image and is superior to a max voting approach. The consideration of contextual information can provide additional assistance in classification tasks within computational pathology (Bejnordi, *et al*. 2017; Agarwalla, *et al*., 2017). For example, growth patterns in LUAD cases and how the tumor grows with the stroma is of significant importance when classifying NSCLC cases. These patterns are often very hard to visualize in a 224×224 patch at 20× resolution. In future work, developing our proposed network to accurately include more contextual information may improve patch-level accuracy and therefore overall classification accuracy. To develop this work, we also aim to use a larger dataset so that our patch-level classifier is able to extract more representative features for subsequent NSCLC classification.

**Acknowledgments**. This work was supported by the National Research Foundation of Korea (NRF) grant funded by the Korean government (MSIP) (No. 2016R1C1B2012433), in part by 1U24CA180924-01A1 from the National Cancer Institute, R01LM011119-01 and R01LM009239 from the U.S. National Library of Medicine.

**Author Contributions Statement.** QV, MT, JK proposed, implemented and evaluated the deep learning method for nucleus segmentation algorithm in Section 2.1. SG, MS, TQ, NK, SK, NR proposed, implemented and evaluated the two-part method for whole slide tissue classification in Section 2.2. KF, TK and JS are the main organizers of the MICCAI 2017 Digital Pathology Challenge. TK, JS, TZ and RG put together the challenge datasets, supervised the generation of ground truth data, and developed the ensemble dice index scoring method. All authors contributed text to the manuscript and edited it.

**Conflict of Interest Statement.** The authors do not have a conflict of interest related to this work.